%% file: emnlp2021.tex
\pdfoutput=1

\documentclass[11pt]{article}

\usepackage{emnlp2021}

\usepackage{graphicx}
\usepackage{eucal}

\usepackage{times}
\usepackage{latexsym}
\usepackage{subfigure}

\usepackage[T1]{fontenc}

\usepackage[utf8]{inputenc}

\usepackage{microtype}

\makeatletter
\newcommand*\bigcdot{\mathpalette\bigcdot@{.8}}
\newcommand*\bigcdot@[2]{\mathbin{\vcenter{\hbox{\scalebox{#2}{$\m@th#1\bullet$}}}}}
\makeatother

\usepackage{algorithm}

\usepackage{multirow}

\title{Switch Point biased Self-Training:\\Re-purposing Pretrained Models for Code-Switching}

\author{
Parul Chopra,
Sai Krishna Rallabandi, 
Alan W Black, Khyathi Raghavi Chandu \\
Language Technologies Institute \\
Carnegie Mellon University \\
\texttt{\{parulcho, srallaba, awb, kchandu\}@andrew.cmu.edu}
}

\begin{document}
\maketitle
\begin{abstract}


\input{sections/abstract}
\end{abstract}



\input{sections/introduction}

\section{Related Work}
\input{sections/literature}

\section{Benchmarking Multilingual Pretrained Models}

\input{sections/methods}



\section{Conclusions}
\input{sections/conclusion}
\bibliography{anthology,custom}
\bibliographystyle{acl_natbib}

\newpage
\pagebreak
\clearpage

\appendix

\section{Experimental Setup}

In the majority of our approaches, we perform task adaptive ﬁne-tuning on BERT, mBERT , XLM-Roberta and character BERT for few  epochs on an Nvidia GeForce GTX 1070 GPU. We primarily used Pytorch and Huggingface library for implementing different models. We experiment with

\noindent $\bigcdot$ batch sizes of 8,16, and 32\\
$\bigcdot$ learning rates between 1e-5 and 5e-5. 

For some models, we observed variation in performance on test set based on subset of data used for training. To overcome this, we did 5 fold cross validation where there were no pre-defined train, dev and test data splits.

\section{Comparing pre-trained Models}
\label{sec:appendix}

Distillation seems to help compared to the corresponding full model. When finetuned for CS cases, distilled variants of BERT and mBERT performed significantly better than their pre-trained counterparts. We plan to investigate the reason behind the mixed results in future work.

\section{Benchmarking Arabic-English}
We observed a similar trend in the benchmarking experiments for Arabic-English code-switching case as well. We performed NER using the dataset by \cite{molina2019overview}. 
These results are shown in Table \ref{tab:benchmarking2}.
We observe that a finetuned BERT model is already much better than the previous state-of-the-art model on the dataset. The M-BERT model further improves this score. However, distilled M-BERT did not show the same improvements as was shown on some other datasets. The trend with distilled models does not seem to be consistent (as discussed in Section B), and we believe that further investigation is needed to understand the reasons behind this performance.
 We do not include this in the results for benchmarking in Table \ref{tab:benchmarking}. 
This is because we could not comprehensively compare the multitasking model with the rest of the models due to the lack of gold label annotations for this dataset (The remaining datasets compared in Table \ref{tab:benchmarking} were annotated with lexical level language ids as well). Finally, while char-BERT showed improvements both over the state of the art model and the finetuned BERT, it did not give the same improvement over the latter. We believe this needs further investigation as well.

\input{tables/appendix_table2}

\section{Self-Training Experiment Details}
We show incremental model performance as we augment training data with batches of un-annotated data in Table \ref{tab:citation-guide1}. As we can observe from the table, the performance of the models increase and then decline after a point when further augmented. We believe the reason behind this is that we are overly biasing the model with this switch point beyond a certain level when the performance starts flipping towards decline. The optimal point of this iterative augmentation with self training is achieved before the flip in the overall performance.

\input{tables/appendix_table1}



\end{document}

%% file: sections/abstract.tex
Code-switching (CS), a ubiquitous phenomenon due to the ease of communication it offers in multilingual communities still remains an understudied problem in language processing. The primary reasons behind this are: 
(1) minimal efforts in leveraging large pretrained multilingual models, and (2) the lack of annotated data. 
The distinguishing case of low performance of multilingual models in CS is the intra-sentence mixing of languages leading to switch points.
We first benchmark two sequence labeling tasks -- POS and NER on 4 different language pairs with a suite of pretrained models to identify the problems and select the best performing model, char-BERT, among them (addressing (1)).
We then propose a self training method to repurpose the existing pretrained models using a switch-point bias by leveraging unannotated data (addressing (2)).
We finally demonstrate that our approach performs well on both tasks by reducing the gap between the switch point performance while retaining the overall performance on two distinct language pairs in both the tasks \footnote{\scriptsize{Our code is available here: \url{https://github.com/PC09/EMNLP2021-Switch-Point-biased-Self-Training}}} . 

%% file: sections/introduction.tex
\section{Introduction}
\label{sec:introduction}

Code-switching (CS) is a phenomenon of switching back and forth between multiple languages and is very common in multilingual communities such as India, Singapore, etc. Understanding mixed language texts has several applications in an increasingly online world like hateful content detection, maintaining engagement with virtual assistants. Despite this pervasive prevalence, CS is often overlooked in language processing research and current models still cannot effectively handle CS. We believe that the reasons behind this are (1) the lack of efforts in leveraging existing large scale multilingual resources or pretrained models and (2) dearth of annotated resources in  switching scenarios. In this paper, we present solutions to address these two problems specifically.

The advent of pretraining techniques marshalled the celebrated successes of several language understanding and generation tasks in English \cite{dong2019unified} and multilingual tasks \cite{chaudhary2020dict}.  However, the same level of commendatory results are not translated to CS scenarios; as studied by
\citet{aguilar-etal-2020-lince, khanuja-etal-2020-gluecos}
presenting a preliminary evaluation of multi-lingual pretrained
models for CS scenarios. 
It is still largely unclear if the inadequacies are resulting due to dearth of data or ineptitude of quick adoption of multilingual models. We study precisely this problem of identifying the artifacts that hinder the competent performance of pretrained models on CS with a case study on sequence labeling tasks including Part-Of-Speech (POS) tagging and Named Entity Recognition (NER).

Our contributions from this work are as follows: (1) We first conduct a comprehensive
benchmarking of different pretrained models 
for two sequence labeling tasks across 4 different language pairs. Specifically we evaluate datasets in Hinglish, Tenglish, Benglish and Spanglish CS for the tasks NER and POS. (2) To broaden understanding towards the usefulness of 
different fine-tuning strategies, we investigate multitasking, character modeling uncovering the problematic switch point cases in
\S \ref{section:SPaugmentation}.
(3) We propose a novel \textit{switch-point bias based self training approach} built upon on observations from the benchmarks and demonstrate improved results on both tasks.

%% file: sections/literature.tex

\paragraph {CS benchmarks:}
From one of the recent surveys \cite{sitaram2019survey}, linguistic CS has been studied in the context of many NLP tasks including language identification \cite{solorio2014overview} \cite{bali2014borrowing}, POS tagging \cite{soto2018joint} \cite{molina2019overview} \cite{das2016tool},  NER \cite{aguilar2019named},  parsing \cite{partanen2018dependency},  sentiment analysis(Vilares  et  al.,  2015),  and  question  answering  \cite{chandu2019code} \cite{raghavi2015answer}.   Many CS datasets have been made available through the shared-task series  FIRE
\cite{choudhury2014overview}; \cite{roy2013overview} and CALCS
\cite{aguilar2018proceedings}, which have focused mostly on core NLP tasks.   Additionally,  other researchers  have  provided  datasets such as humor detection \cite{khandelwal2018humor},  sub-word  CS  detection  \cite{mager2019subword} among others. More recently new CS benchmarks \cite{aguilar-etal-2020-lince} \cite{khanuja-etal-2020-gluecos} have been developed to compare models across language pairs, domains and general language processing in CS.


\paragraph{Pretrained Models for CS:}
Before the advent of pretrained multingual models, pretrained monolingual models were combined in different ways to derive word embeddings
\cite{alghamdi2019leveraging, pratapa2018word}, POS tagging \cite{bhattu2020improving}, sentiment analysis \cite{singh2020sentiment} etc.,
Similarly, pretrained multilingual models have been explored on various CS tasks like language identification, POS tagging, NER, question answering and Natural language inference \cite{khanuja-etal-2020-gluecos}. 
However, \cite{winata2021multilingual} show that these pretrained models do not assure high quality representations on CS.
We examine prospective reasons for this and present a data augmentation technique to mitigate this.


\paragraph{Motivation for our work - Gaps in CS adaptation: }
Building off the prior work, we will briefly discuss primarily three techniques that demonstrated usefulness in adapting models to CS.
First, non-standardization of cross-scripting (i.e, transliteration of words to another language) is identified as one of the major reasons behind the noisiness of CS datasets \cite{chandu2019code}.
Prior literature on noisy texts proved the superiority of character level modeling to combat this problem \cite{cherry2018revisiting}; \cite{adouane2018comparison}. Secondly, the domains of most of these noisy datasets are still vastly scattered. In order to improve generalization in CS patterns, prior studies have shown the potency of multitasking with an auxiliary task of language tag prediction \cite{winata2018code}. Thirdly, the dearth of annotated CS data has been a dramatic problem across tasks. \cite{bhattu2020improving} compare pretrained models with fined-tuned models augmented with unlabeled Twitter text to exemplify the improved performance with the latter model.
Despite these takeaways, the usefulness of the three points above is not thoroughly investigated in the context of pretrained  models for CS. To this end, we adapt these techniques in conjunction with the pretraining strategies and propose a novel bias-based data iterative augmentation technique to get more bang for the buck in terms of the performance to augmented dataset size ratio.





%% file: sections/methods.tex
\subsection{Datasets and Models}

\input{tables/datasets}

\input{tables/pos}

We selected datasets from LinCE\cite{aguilar-etal-2020-lince} and GlueCOS \cite{khanuja-etal-2020-gluecos} benchmarks for all our experiments. The details of these datasets are presented in Table \ref{tab:citation}.  We present a comprehensive evaluation of different 
BERT-based mono-lingual and multi-lingual pretrained models
when adapted to the chosen CS datasets/tasks. 
We performed sequence tagging on different transformer models: (a) We use the uncased base implementation of \textbf{BERT}  and \textbf{mBERT} \cite{devlin2018bert}  (b) \textbf{Distill mBERT} \cite{sanh2019distilbert}, (c)  \textbf{XLM-RoBERTa} \cite{conneau2019unsupervised} trained using knowledge distillation and (d) \textbf{Char-BERT}\cite{boukkouri2020characterbert} that employs Character CNN to capture unknown and misspelled words. Motivated by prior works on multi-task learning \cite{chandu2018language, li2020low}, we also experiment with language-aware modeling. In these experiments, we added a language token either as the \textit{input encoding} or \textit{output prediction}.

\subsection{Analysis of Benchmarking}
\label{sec:analysis}
The results for the aforementioned experiments are presented in Table \ref{tab:benchmarking}. The baseline in this table indicates the  current state-of-the-art models on respective datasets as cited in the table. Here are our main observations from these results.



\vspace{0.2cm}

\noindent $\bigcdot$ \textit{Multi-task Learning did not help much: } Despite the effectiveness of multi-tasking in non-pretrained (models trained from scratch) CS modeling, vast improvements are not observed upon finetuning pretrained with multitasking objective. 

\vspace{0.2cm}

\noindent $\bigcdot$ \textit{Improvement with Char-BERT: } We observe that Char-BERT gives significant improvement in POS specifically for Indic sets: English-Bengali and English-Telugu. On others, its performance is comparable to current SOTA with mBERT or XLM-RoBERTa. Although the languages in the pretraining of mBERT include the language pairs of concern here, we do not observe benefits from this model as the training data mostly includes data from the script of the source language. For example, 
training on Devanagari Hindi does not necessarily translate its ability to understand the cross-scripted and usually Romanized CS texts.
\vspace{0.2cm}

\begin{figure}[t!]
  \centering
  \subfigure[SP count ]
  {\includegraphics[scale=0.27]{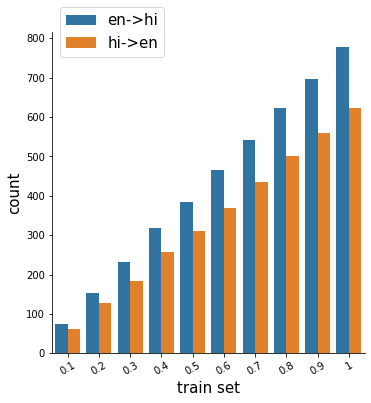}}\quad
  \subfigure[SP validation accuracy]
  {\includegraphics[scale=0.27]{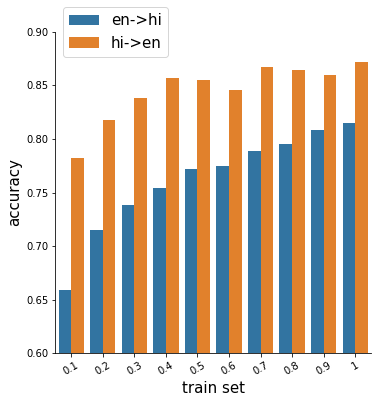}}
 
  \caption{(a) Count and (b) accuracy over val set for different portions or percentages of training data (\textit {EnHi-Tw-P)} for both switch-points (SP)}
\label{fig:count-acc}
\end{figure}

\noindent $\bigcdot$ \textit{Performance at switch-points: } We further investigated the performance at switch-points which distinguishes CS from monolingual texts. We demonstrate this for \textit{EnHi-Tw-P} in Figure \ref{fig:count-acc}, where the validation accuracy of switching from English to Hindi (en $\rightarrow$ hi) is relatively much lower compared to switching from Hindi to English (hi $\rightarrow$ en). We observe this pattern to be consistent across the datasets in Table \ref{tab:citation-guide} and propose a solution to address this in the next section.

\section{Switch-Point biased Self Training}
\label{section:SPaugmentation}

As observed in the previous section, performance of the models deteriorates at the switching points \cite{chatterjere2020minority} in CS. This motivates our approach to tackle this problem which can be stated concretely as: 
\begin{quote}
\small
\textit{The pre-trained model favors embedded-to-matrix over matrix-to-embedded language switching points despite majority of training data in the former pattern.} 
\end{quote}

We demonstrate this by comparing Figure \ref{fig:count-acc}(a) and Figure \ref{fig:count-acc}(b) for the case of \textit{EnHi-Tw-P}. They present the counts and val accuracy with the increasing percentage of train set on the x-axis to demonstrate the consistency of this pattern. As we can observe, the number of samples with switch points from en $\rightarrow$ hi is higher than that of hi $\rightarrow$ en (Fig \ref{fig:count-acc}(a)). However, the performance on switch points from en $\rightarrow$ hi is relatively much lower than the counter part (Fig \ref{fig:count-acc}(b)).

\begin{algorithm}[H]
\small
\textbf{Input: } Annotator Model A($\theta$), Labeled Data $\CMcal{D}^{l}$, Unlabeled Data $\CMcal{D}^{u}$ \\
\textbf{Output: } Trained End-Task Model $\CMcal{E}(\phi^{'}$) \\
1. Fine-tune $\CMcal{E}(\phi)$ on $\CMcal{D}^{l}$ \\
2. s $\leftarrow$ Identify the low-performing switch-point \\
3. $\CMcal{D}^{u}_{s}$ $\leftarrow$ Sub-sample data from $\CMcal{D}^{u}$ with higher ratio of s \\
4. $\CMcal{D}^{wl}_{s}$ $\leftarrow$ Annotate $\CMcal{D}^{u}_{s}$ with $\CMcal{A}(\theta^{'}$)  \\
5. $\CMcal{A}$($\theta^{'}$) $\leftarrow$ Fine-tune $\CMcal{A}(\theta)$ on $\CMcal{D}^{wl}_{s}$ + $\CMcal{D}^{l}$ \\
6. $\CMcal{E}(\phi^{'})$ $\leftarrow$ Train $\CMcal{E}(\phi)$ on $\CMcal{D}^{wl}_{s}$ + $\CMcal{D}^{l}$ \\
7. Repeat Steps 2 to 6 by updating  $\CMcal{A}(\theta^{'})$ and $\CMcal{E}(\phi^{'})$ \\
 \caption{\small Switch-Point biased Self Training (selfTr) }
 \label{alg:fst}
\end{algorithm}

\input{tables/augmentation}

\noindent We posit that a switch point specific fine-tuning is required to combat this imbalance. Our proposed approach is depicted in Algorithm \ref{alg:fst}. The baseline for each task is the char-BERT model fine-tuned on the task-specific data, which is referred as end-task model $\CMcal{E}(\phi)$.

Our first step is to compute switch point ratios. We computed the percentage of switch points from En->X  (say a) and from X->En (say b) on the unlabeled data. We then compute s=a/b. If s<1, we bias our annotator model by training with the sentences that has `s greater than 1' i.e biased to En->X data, otherwise, we train it with the sentences that has `s lesser than 1' i.e biased to X->En data. In this way, our annotator model is biased to favor annotations on low-performing switch-point and is further used to annotate the unlabeled dataset. 

We then identify the low-performing switch point and derive the Annotator Model A($\theta$) with the labeled subset of the low-performing switch point (\textit{s}) from the dataset. This annotator model is now biased to favor annotations on this \textit{s} to increase its bias for further annotations. We leverage a vast amount of unlabeled dataset  $\CMcal{D}^{u}$. The unlabeled data is gathered from the validation and test subsets of the standard datasets (from Table \ref{tab:citation}) without considering the true labels. We use the raw samples i.e., sentences and annotate them using the annotator model. Based on the amount of samples available, we iteratively annotate and add samples to our original training dataset with our switch-point bias based self training.

The underlying annotator model can be any of the large scale pretrained models that we experimented with in the previous section. We choose to use char-BERT as our annotator model. This annotator model is used to annotate the subset of the unlabeled data with sequence tags. This weakly annotated noisy data is now augmented to the labeled dataset. Both the annotator model and the end-task model are now finetuned with this augmented dataset. This iterative data augmentation process repeats until the performance stops degrading.

\begin{figure*}[t!]
\centering
\includegraphics[trim=1.2cm 6.2cm 2cm 3cm, clip,width=0.8\linewidth]{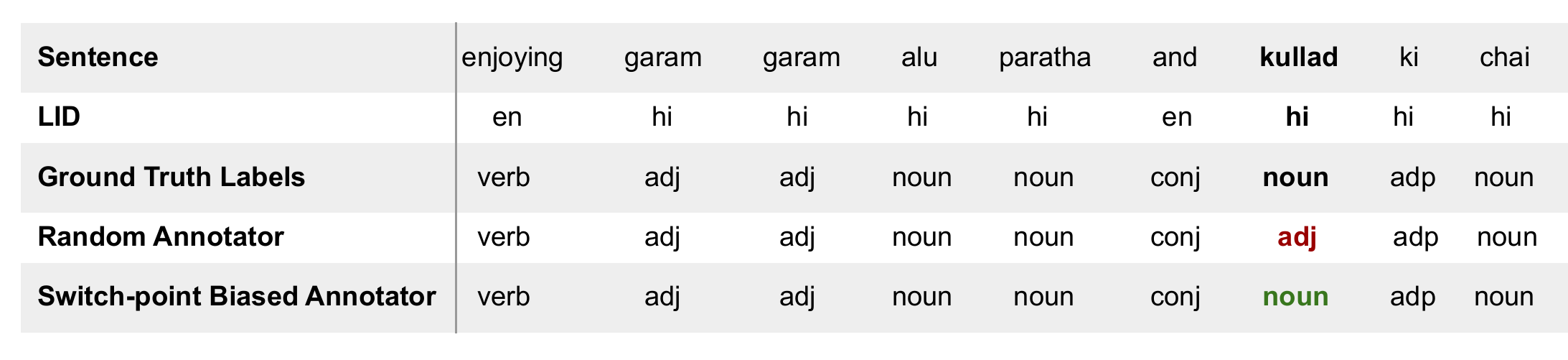}
\caption{\small {Example of model predictions from Random Annotator model and Switch-point biased Annotator model. \textit{(Meaning of the example sentence: Enjoying hot potato bread and kullad tea.)}} }
\label{fig:ea}
\end{figure*}

\subsection{Results}
Adding the annotated data via switch point based self training helps the model better learn at low-performing code switching points. In Table \ref{tab:citation-guide}, we evaluate this technique on 4 different datasets where we train both our model and annotator by fine-tuning a character-BERT model (as we observed improvements with this model in Section \ref{sec:analysis}). Note that X refers to the language which is mixed with English.
We can see that among the char-BERT baseline (first row in each segment of the table), the performance is highly biased both in terms of F1 and accuracy towards: (i) switching to English (X $\rightarrow$ en switch point) in the first 3 segments, and (ii) switching to Spanish (en $\rightarrow$ X switch point) in the last segment.
Accordingly we train annotator models described above and augment the training data. 
To evaluate the effectiveness of our approach, we also compare these results to the case when annotator model is updated by training with augmented data selected randomly of the same size. It can be seen that our bias based approach performs better than uninformed random data augmentation for training.
Our approach demonstrates consistent improvements at the low-performing switch points. 
The difference in switch-point F1 scores between X $\rightarrow$ en F1 and en $\rightarrow$ X F1 compared between the baseline char-BERT and our approach is reduced by a margin of 5\%, 3\%, 6\% and 5\% on POS English-Hindi \cite{singh2018twitter}, POS English-Spanish \cite{alghamdi2019part}, NER English-Hindi \cite{singh2018named}  and NER English-Spanish \cite{aguilar2019named} respectively. 
In this way, we also improved the overall accuracy and F1 in 3 and 2 datasets respectively, while the scores remained almost the same for 1 and 2 datasets correspondingly.

Figure \ref{fig:ea} presents an example sentence from Hindi-English code-switched POS data along with language ids along with ground truth labels and predictions. The random annotator model incorrectly predicts `\textit{kullad}' as \textit{adj} when transitioning from English to Hindi (en $\rightarrow$ X). Our switch-point biased based model correctly labels this word. 

\paragraph{Analysis: } An inspection of pretrained models revealed different types of errors: (a) Errors on \textbf{NUM} when the numerals were in Hindi and (b) Confusion between the classes \textbf{PROPN} and \textbf{N} (c) Errors due to misspelled words and (d) logical errors due to ambiguous sentences. In general we observed some noise in the dataset labels itself.

We also conducted a categorical error analysis of the performance on one of the language pairs that is Hindi-English data. In this language mixing, for example, we noticed that when switching from X $\rightarrow$ En, the errors are significantly higher for Proper Nouns ($\sim$99\%) and Interjections ($\sim$99\%) as compared to other POS tags, while the reverse is the case for Determiners ($\sim$98\%) and Particles ($\sim$94\%). The numbers in the brackets indicate the `absolute difference' of accuracies between En $\rightarrow$ X and X $\rightarrow$ En for predictions of the corresponding POS tag. This means that Proper Nouns and Interjections are more difficult to tag when switched from Hindi to English, but the same pattern is not observed when switched from English to Hindi.





  



%% file: tables/datasets.tex
\begin{table}
\scriptsize
\centering
\begin{tabular}{llllll}
\hline
\textbf{Corpus} & \textbf{Notation} & \textbf{Task} & \textbf{\# Sentences} \\
\hline


Twitter \cite{singh2018twitter}  & \textbf{EnHi-Tw-P} & POS  & 1489 \\

UD \cite{bhat2018universal}  & \textbf{EnHi-UD-P} & POS & 1311 \\

ICON \cite{jamatia2016collecting}  & \textbf{EnHi-I-P} & POS & 2630\\


ICON \cite{jamatia2016collecting} & \textbf{EnBn-I-P}  & POS  & 625  \\

ICON \cite{jamatia2016collecting}  & \textbf{EnTe-I-P} & POS & 1979 \\

Miami \cite{alghamdi2019part}  & \textbf{EnEs-M-P} & POS & 27893 \\ \hline

Twitter \cite{singh2018named}  & \textbf{EnHi-Tw-N} & NER & 1243  \\


CALCS \cite{aguilar2019named}   & \textbf{EnEs-Tw-N} & NER & 50757  \\


\hline
\end{tabular}
\caption{\label{tab:citation} Details of CS datasets \& training sizes}

\end{table}

%% file: tables/pos.tex
\begin{table*}
\scriptsize
\centering
\begin{tabular}{l|llllll|ll}

\hline
\multirow{1}{*}{\textbf{Model}} & \multicolumn{6}{|c|}{\textbf{Part-Of-Speech tagging}} & \multicolumn{2}{c}{\textbf{Named Entity Recognition}} \\

\hline
\multicolumn{1}{c|}{\textbf{}} & 
\multicolumn{1}{c}{\textbf{EnHi-Tw-P}} &  \multicolumn{1}{c}{\textbf{EnHi-UD-P}} &
\multicolumn{1}{c}{\textbf{EnEs-M-P}} &
\multicolumn{1}{c}{\textbf{EnHi-I-P}} &
\multicolumn{1}{c}{\textbf{EnBn-I-P}} &
\multicolumn{1}{c|}{\textbf{EnTe-I-P}} &
\multicolumn{1}{c}{\textbf{EnHi-Tw-N}} &
\multicolumn{1}{c}{\textbf{EnEs-Tw-N}} \\


\hline
Baseline & 91.03 (A) & 90.53 (B) &95.39(B) & 85.26 (C) & 77.15 (C) & 74.88 (C) & 78.21 (B)  & 69.17 (B) \\

eng-BERT & 84.01 & 82.12 & 91.77 & 80.55 & 75.78 & 76.11 & 65.93 & 55.12 \\
M-BERT & 89.27 & 87.67 & 93.12 & 86.38 & 80.74 & 79.01  & 74.2 & 60.12 \\


M-BERT (lang-input) & 89.74 & 87.96 & 93.65 & 86.99 & 81.67 & 78.55 & 75.38 & 61.46\\
M-BERT (lang-output) & 88.89 & 86.47 & 92.89 & 85.65 & 81.17 & 76.13 & 74.01 & 60.20\\

Distill M-BERT & 90.28 & 88.19 & 93.65  & 86.92 & 82.07 & 79.85 &  67.26 & 62.67 \\
XLM-ROBERTa & 90.74 & 89.88 & 95.34 & 86.24 & 80.58 & 75.83 & 73.34 & 66.12\\

char-BERT & 90.89 & 90.23 &96.88 & 87.11 & 82.21 & 80.33 & 77.24 & 65.72 \\

char-BERT (lang-input) & 91.02 & 90.93  &97.01 & 87.24 & 82.87 & 82.52 & 77.43 & 66.34 \\
char-BERT (lang-output) & 90.25 & 89.29 &96.25 & 86.39 & 82.47 & 80.98 & 77.12 & 66.01 \\

\hline
\end{tabular}
\caption{\label{tab:benchmarking}
Performance of different multilingual models for various POS tagging datasets (Accuracy), NER (F1) in single , multi-task setting (language at input/output). Results are reported for datasets- 
(A) \cite{aguilar2020english},
(B) \cite{khanuja-etal-2020-gluecos},
(C) \cite{bhattu2020improving}
}
\end{table*}

%% file: tables/augmentation.tex
\begin{table*}[t!]
\small
\centering
\begin{tabular}{l|l|ll|ll|ll}
\hline

\hline 
\multicolumn{1}{c|}{} & 
\multicolumn{1}{c|}{\textbf{Biased}} & 
\multicolumn{2}{c|}{\textbf{Overall}} & 
\multicolumn{2}{c|}{\textbf{X $\rightarrow$ en}} &
\multicolumn{2}{c}{\textbf{en $\rightarrow$ X}} \\ 

\textbf{Model} & \textbf{Annotator}& \textbf{Acc.} & \textbf{F1} & \textbf{Acc.} & \textbf{F1} & \textbf{Acc.} & \textbf{F1}\\

\hline

POS-en-hi & - & 89.84  & 85.21 & 87.91 & 86.11 & 82.62 & 80.23\\
POS -en-hi-random & - & 89.29  & 85.11 & 88.78 & 87.23 & 82.40 & 80.11\\

POS-en-hi-selfTr & en $\rightarrow$ hi & \textbf{89.91}  & 85.16 & 87.27 & 86.34 & \textbf{84.38} & \textbf{85.01}\\ \hline

POS-en-es & - & 96.88 & 96.25  & 93.82 & 90.97 & 88.59 & 85.38\\
POS -en-es-random & - & 96.91 & 96.21  & 94.10 & 90.12 & 88.04 & 84.32\\

POS-en-es-selfTr & en $\rightarrow$ es & \textbf{97.05} & \textbf{96.41} & \textbf{95.51} & \textbf{93.42} & \textbf{90.6} & \textbf{88.12}\\ \hline

NER-en-hi & - & 95.45 & 75.18  & 96.92 & 84.84 & 93.21 & 77.41\\ 
NER-en-hi-random & - & 95.42 & 75.12   & 97.05 & 86.12 & 93.09 & 76.65\\ 
NER-en-hi-selfTr & en $\rightarrow$ hi & 95.41  & 75.02 & 95.89 & 78.78 & \textbf{95.02} & \textbf{80.70} \\ \hline

NER-en-es & -& 93.00 & 65.72  & 83.28 & 47.73 & 93.35 & 58.19 \\
NER-en-es-random & - & 93.10 & 65.95 & 84.25  & 49.22 & 94.10 & 59.67  \\

NER-en-es-selfTr & es $\rightarrow$ en & \textbf{93.12} & \textbf{66.34} & \textbf{86.13} & \textbf{56.62} & \textbf{94.58} & \textbf{62.29} \\
\hline
\end{tabular}
\caption{\label{tab:citation-guide}
Results of our \textit{switch point biased self training} (selfTr). Here the annotator model is trained on subset of data which is more biased towards lower-performing switch point. The biased annotator model is trained using a subset of the data with the switch point shown in the table. X refers to the language which is mixed with English.
}

\end{table*}

%% file: sections/conclusion.tex
CS, despite being a natural and prevalent form of communication is still vastly understudied in empirical research. This mainly stems from the (1) lack of efforts in re-purposing the celebrated pretrained models to CS scenarios and (2) lack of annotated resources. We tackle precisely these 2 problems with the main focus on evaluating and improving how these models fare at switch points between languages. First, we benchmark a suite of monolingual and multilingual pretrained models on CS and identify that particular switch points fare poorly.
We propose a novel switch point bias based self training method to strategically use unlabeled data to enhance performance at switch points. While improving or retaining the overall performance compared to finetuning char-BERT and multitasking, we show that our approach improves the performance of underperforming switch points as well. We believe that this bias based augmentation technique particularly helps in scenarios with less annotated data.

\section{Broader Impact}
\label{sec:impact}
We believe that this work is a step towards effacing the hesitation of utilizing large scale pretrained mono and multilingual models for code-switched scenarios. We were able to successfully demonstrate the utility of a switch point based annotator model to perform biased data augmentation. We do not foresee any immediate ethical concerns branching directly from our work. However, we cautiously advise anyone using or extending our work for their application or research to bear in mind that we inherit any kinds of biases and toxicity and privacy concerns that the pretrained language models bear. Although our end tasks are not directly affected forthwith due to these, we still recommend caution when our self training approach is used for other tasks especially with user interaction such as dialog response generation etc., to ensure the model does not predict toxic content.
Overall, we expect the users to benefit from our research to prospectively apply this to scenarios where there is a dearth of annotated resources, thereby economizing on annotations cost and efforts and enabling scaling up to a wealth of crawled data, if available in those language-pairs.

%% file: tables/appendix_table2.tex
\begin{table}[tbh]
\small
\centering
\begin{tabular}{l|llllll|ll}

\hline
\multirow{1}{*}{\textbf{Model}} & \multicolumn{6}{|c}{\textbf{Named Entity Recognition}} \\

\hline
\multicolumn{1}{c|}{\textbf{}} & 
\multicolumn{1}{c}{\textbf{msa-ea-N}} \\

\hline
Baseline & 71.61 \\

eng-BERT & 74.13 \\

M-BERT & 79.73  \\

Distill M-BERT & 77.28  \\

XLM-ROBERTa & 77.68\\

char-BERT & 74.46   \\

\hline
\end{tabular}
\caption{\label{tab:benchmarking2}
Performance of different multilingual models on MSA-EA \cite{molina2019overview} dataset. 
}
\end{table}

%% file: tables/appendix_table1.tex
\begin{table*}[tbh]
\small
\centering
\begin{tabular}{l|l|l|ll|ll|ll}
\hline

\hline 
\multicolumn{1}{c|}{} & 
\multicolumn{1}{c|}{\textbf{Biased}} & 
\multicolumn{1}{c|}{\textbf{Sentences}} & 
\multicolumn{2}{c|}{\textbf{Overall}} & 
\multicolumn{2}{c|}{\textbf{X $\rightarrow$ en}} &
\multicolumn{2}{c}{\textbf{en $\rightarrow$ X}} \\ 

\textbf{Model} & \textbf{Annotator} & \textbf{Added} & \textbf{Acc.} & \textbf{F1} & \textbf{Acc.} & \textbf{F1} & \textbf{Acc.} & \textbf{F1}\\

\hline

POS-en-hi & - & - &  89.84  & 85.21 & 87.91 & 86.11 & 82.62 & 80.23\\

POS-en-hi-selfTr & en $\rightarrow$ hi & +400 &  89.89 & 85.32 & 87.727 &86.23 & 83.13 & 82.11\\ 

POS-en-hi-selfTr & en $\rightarrow$ hi & +400 &  89.91 & 85.16 & 87.27 &86.34 & 84.38& 85.01\\ 

\hline

POS-en-es & -& - & 96.88 & 96.25  & 93.82 & 90.97 & 88.59 & 85.38\\
POS-en-es-selfTr & en $\rightarrow$ es & +150 & 97.01 & 96.29 & 94.15 & 91.02 & 89.66 & 87.01\\ 
POS-en-es-selfTr & en $\rightarrow$ es & +150 & 97.05 & 96.41 & 95.51 & 93.42 & 90.6 & 88.12\\ 

\hline

NER-en-hi & -& -  & 95.45 & 75.18  & 96.92 & 84.84 & 93.21 & 77.41\\ 

NER-en-hi-selfTr & en $\rightarrow$ hi & +100 & 95.71  & 77.44 & 96.92 & 83.11 & 93.66 & 77.96 \\

NER-en-hi-selfTr & en $\rightarrow$ hi & +100 & 95.57  & 77.01 & 96.41 & 83.87 & 94.57 & 80.70 \\

NER-en-hi-selfTr & en $\rightarrow$ hi & +100 & 95.41  & 75.02 & 95.89 & 78.78 & 95.02 & 80.70 \\

\hline

NER-en-es & - & - &  93.00 & 65.72  & 83.28 & 47.73 & 93.35 & 58.19 \\


NER-en-es-selfTr & es $\rightarrow$ en & +500 & 93.32 & 65.84 & 83.89 & 50.62 & 93.98 & 60.29 \\

NER-en-es-selfTr & es $\rightarrow$ en & +500 & 93.43 & 66.14 & 84.75 & 53.54 & 93.5 & 60.89 \\

NER-en-es-selfTr & es $\rightarrow$ en & +500 & 93.12 & 66.34 & 86.13 & 56.62 & 94.58 & 62.29 \\

\hline
\end{tabular}
\caption{\label{tab:citation-guide1}
Results from Switch point biased self training. X refers to the language which is mixed with English. Iteratively \# number of sentences are added to training set.
}

\end{table*}

%% file: emnlp2021.bbl
\begin{thebibliography}{40}
\expandafter\ifx\csname natexlab\endcsname\relax\def\natexlab#1{#1}\fi

\bibitem[{Adouane et~al.(2018)Adouane, Dobnik, Bernardy, and
  Semmar}]{adouane2018comparison}
Wafia Adouane, Simon Dobnik, Jean-Philippe Bernardy, and Nasredine Semmar.
  2018.
\newblock A comparison of character neural language model and bootstrapping for
  language identification in multilingual noisy texts.
\newblock In \emph{Proceedings of the Second Workshop on Subword/Character
  LEvel Models}, pages 22--31.

\bibitem[{Aguilar et~al.(2019)Aguilar, AlGhamdi, Soto, Diab, Hirschberg, and
  Solorio}]{aguilar2019named}
Gustavo Aguilar, Fahad AlGhamdi, Victor Soto, Mona Diab, Julia Hirschberg, and
  Thamar Solorio. 2019.
\newblock Named entity recognition on code-switched data: Overview of the calcs
  2018 shared task.
\newblock \emph{arXiv preprint arXiv:1906.04138}.

\bibitem[{Aguilar et~al.(2018)Aguilar, AlGhamdi, Soto, Solorio, Diab, and
  Hirschberg}]{aguilar2018proceedings}
Gustavo Aguilar, Fahad AlGhamdi, Victor Soto, Thamar Solorio, Mona Diab, and
  Julia Hirschberg. 2018.
\newblock Proceedings of the third workshop on computational approaches to
  linguistic code-switching.
\newblock In \emph{Proceedings of the Third Workshop on Computational
  Approaches to Linguistic Code-Switching}.

\bibitem[{Aguilar et~al.(2020)Aguilar, Kar, and
  Solorio}]{aguilar-etal-2020-lince}
Gustavo Aguilar, Sudipta Kar, and Thamar Solorio. 2020.
\newblock \href {https://www.aclweb.org/anthology/2020.lrec-1.223} {{L}in{CE}:
  A centralized benchmark for linguistic code-switching evaluation}.
\newblock In \emph{Proceedings of the 12th Language Resources and Evaluation
  Conference}, pages 1803--1813, Marseille, France. European Language Resources
  Association.

\bibitem[{Aguilar and Solorio(2020)}]{aguilar2020english}
Gustavo Aguilar and Thamar Solorio. 2020.
\newblock \href {http://arxiv.org/abs/1909.05158} {From english to
  code-switching: Transfer learning with strong morphological clues}.

\bibitem[{AlGhamdi and Diab(2019)}]{alghamdi2019leveraging}
Fahad AlGhamdi and Mona Diab. 2019.
\newblock Leveraging pretrained word embeddings for part-of-speech tagging of
  code switching data.
\newblock \emph{arXiv preprint arXiv:1905.13359}.

\bibitem[{AlGhamdi et~al.(2019)AlGhamdi, Molina, Diab, Solorio, Hawwari, Soto,
  and Hirschberg}]{alghamdi2019part}
Fahad AlGhamdi, Giovanni Molina, Mona Diab, Thamar Solorio, Abdelati Hawwari,
  Victor Soto, and Julia Hirschberg. 2019.
\newblock Part of speech tagging for code switched data.
\newblock \emph{arXiv preprint arXiv:1909.13006}.

\bibitem[{Bali et~al.(2014)Bali, Sharma, Choudhury, and
  Vyas}]{bali2014borrowing}
Kalika Bali, Jatin Sharma, Monojit Choudhury, and Yogarshi Vyas. 2014.
\newblock “i am borrowing ya mixing?" an analysis of english-hindi code
  mixing in facebook.
\newblock In \emph{Proceedings of the First Workshop on Computational
  Approaches to Code Switching}, pages 116--126.

\bibitem[{Bhat et~al.(2018)Bhat, Bhat, Shrivastava, and
  Sharma}]{bhat2018universal}
Irshad~Ahmad Bhat, Riyaz~Ahmad Bhat, Manish Shrivastava, and Dipti~Misra
  Sharma. 2018.
\newblock Universal dependency parsing for hindi-english code-switching.
\newblock \emph{arXiv preprint arXiv:1804.05868}.

\bibitem[{Bhattu et~al.(2020)Bhattu, Nunna, Somayajulu, and
  Pradhan}]{bhattu2020improving}
S~Nagesh Bhattu, Satya~Krishna Nunna, DVLN Somayajulu, and Binay Pradhan. 2020.
\newblock Improving code-mixed pos tagging using code-mixed embeddings.
\newblock \emph{ACM Transactions on Asian and Low-Resource Language Information
  Processing (TALLIP)}, 19(4):1--31.

\bibitem[{Boukkouri et~al.(2020)Boukkouri, Ferret, Lavergne, Noji, Zweigenbaum,
  and Tsujii}]{boukkouri2020characterbert}
Hicham~El Boukkouri, Olivier Ferret, Thomas Lavergne, Hiroshi Noji, Pierre
  Zweigenbaum, and Junichi Tsujii. 2020.
\newblock Characterbert: Reconciling elmo and bert for word-level
  open-vocabulary representations from characters.
\newblock \emph{arXiv preprint arXiv:2010.10392}.

\bibitem[{Chandu et~al.(2019)Chandu, Loginova, Gupta, Genabith, Neumann,
  Chinnakotla, Nyberg, and Black}]{chandu2019code}
Khyathi Chandu, Ekaterina Loginova, Vishal Gupta, Josef~van Genabith,
  G{\"u}nter Neumann, Manoj Chinnakotla, Eric Nyberg, and Alan~W Black. 2019.
\newblock Code-mixed question answering challenge: Crowd-sourcing data and
  techniques.
\newblock In \emph{Third Workshop on Computational Approaches to Linguistic
  Code-Switching}, pages 29--38. Association for Computational Linguistics
  (ACL).

\bibitem[{Chandu et~al.(2018)Chandu, Manzini, Singh, and
  Black}]{chandu2018language}
Khyathi Chandu, Thomas Manzini, Sumeet Singh, and Alan~W Black. 2018.
\newblock Language informed modeling of code-switched text.
\newblock In \emph{Proceedings of the Third Workshop on Computational
  Approaches to Linguistic Code-Switching}, pages 92--97.

\bibitem[{Chatterjere et~al.(2020)Chatterjere, Guptha, Chopra, and
  Das}]{chatterjere2020minority}
Arindam Chatterjere, Vineeth Guptha, Parul Chopra, and Amitava Das. 2020.
\newblock Minority positive sampling for switching points-an anecdote for the
  code-mixing language modeling.
\newblock In \emph{Proceedings of The 12th Language Resources and Evaluation
  Conference}, pages 6228--6236.

\bibitem[{Chaudhary et~al.(2020)Chaudhary, Raman, Srinivasan, and
  Chen}]{chaudhary2020dict}
Aditi Chaudhary, Karthik Raman, Krishna Srinivasan, and Jiecao Chen. 2020.
\newblock Dict-mlm: Improved multilingual pre-training using bilingual
  dictionaries.
\newblock \emph{arXiv preprint arXiv:2010.12566}.

\bibitem[{Cherry et~al.(2018)Cherry, Foster, Bapna, Firat, and
  Macherey}]{cherry2018revisiting}
Colin Cherry, George Foster, Ankur Bapna, Orhan Firat, and Wolfgang Macherey.
  2018.
\newblock Revisiting character-based neural machine translation with capacity
  and compression.
\newblock \emph{arXiv preprint arXiv:1808.09943}.

\bibitem[{Choudhury et~al.(2014)Choudhury, Chittaranjan, Gupta, and
  Das}]{choudhury2014overview}
Monojit Choudhury, Gokul Chittaranjan, Parth Gupta, and Amitava Das. 2014.
\newblock Overview of fire 2014 track on transliterated search.
\newblock \emph{Proceedings of FIRE}, pages 68--89.

\bibitem[{Conneau et~al.(2019)Conneau, Khandelwal, Goyal, Chaudhary, Wenzek,
  Guzm{\'a}n, Grave, Ott, Zettlemoyer, and Stoyanov}]{conneau2019unsupervised}
Alexis Conneau, Kartikay Khandelwal, Naman Goyal, Vishrav Chaudhary, Guillaume
  Wenzek, Francisco Guzm{\'a}n, Edouard Grave, Myle Ott, Luke Zettlemoyer, and
  Veselin Stoyanov. 2019.
\newblock Unsupervised cross-lingual representation learning at scale.
\newblock \emph{arXiv preprint arXiv:1911.02116}.

\bibitem[{Das(2016)}]{das2016tool}
Amitava Das. 2016.
\newblock Tool contest on pos tagging for code-mixed indian social media
  (facebook, twitter, and whatsapp) text.

\bibitem[{Devlin et~al.(2018)Devlin, Chang, Lee, and
  Toutanova}]{devlin2018bert}
Jacob Devlin, Ming-Wei Chang, Kenton Lee, and Kristina Toutanova. 2018.
\newblock Bert: Pre-training of deep bidirectional transformers for language
  understanding.
\newblock \emph{arXiv preprint arXiv:1810.04805}.

\bibitem[{Dong et~al.(2019)Dong, Yang, Wang, Wei, Liu, Wang, Gao, Zhou, and
  Hon}]{dong2019unified}
Li~Dong, Nan Yang, Wenhui Wang, Furu Wei, Xiaodong Liu, Yu~Wang, Jianfeng Gao,
  Ming Zhou, and Hsiao-Wuen Hon. 2019.
\newblock Unified language model pre-training for natural language
  understanding and generation.
\newblock \emph{arXiv preprint arXiv:1905.03197}.

\bibitem[{Jamatia et~al.(2016)Jamatia, Gamb{\"a}ck, and
  Das}]{jamatia2016collecting}
Anupam Jamatia, Bj{\"o}rn Gamb{\"a}ck, and Amitava Das. 2016.
\newblock Collecting and annotating indian social media code-mixed corpora.
\newblock In \emph{International Conference on Intelligent Text Processing and
  Computational Linguistics}, pages 406--417. Springer.

\bibitem[{Khandelwal et~al.(2018)Khandelwal, Swami, Akhtar, and
  Shrivastava}]{khandelwal2018humor}
Ankush Khandelwal, Sahil Swami, Syed~S Akhtar, and Manish Shrivastava. 2018.
\newblock Humor detection in english-hindi code-mixed social media content:
  Corpus and baseline system.
\newblock \emph{arXiv preprint arXiv:1806.05513}.

\bibitem[{Khanuja et~al.(2020)Khanuja, Dandapat, Srinivasan, Sitaram, and
  Choudhury}]{khanuja-etal-2020-gluecos}
Simran Khanuja, Sandipan Dandapat, Anirudh Srinivasan, Sunayana Sitaram, and
  Monojit Choudhury. 2020.
\newblock \href {https://doi.org/10.18653/v1/2020.acl-main.329} {{GLUEC}o{S}:
  An evaluation benchmark for code-switched {NLP}}.
\newblock In \emph{Proceedings of the 58th Annual Meeting of the Association
  for Computational Linguistics}, pages 3575--3585, Online. Association for
  Computational Linguistics.

\bibitem[{Li et~al.(2020)Li, Li, Sheng, and Slamu}]{li2020low}
Xiuhong Li, Zhe Li, Jiabao Sheng, and Wushour Slamu. 2020.
\newblock Low-resource text classification via cross-lingual language model
  fine-tuning.
\newblock In \emph{China National Conference on Chinese Computational
  Linguistics}, pages 231--246. Springer.

\bibitem[{Mager et~al.(2019)Mager, {\c{C}}etino{\u{g}}lu, and
  Kann}]{mager2019subword}
Manuel Mager, {\"O}zlem {\c{C}}etino{\u{g}}lu, and Katharina Kann. 2019.
\newblock Subword-level language identification for intra-word code-switching.
\newblock \emph{arXiv preprint arXiv:1904.01989}.

\bibitem[{Molina et~al.(2019)Molina, AlGhamdi, Ghoneim, Hawwari,
  Rey-Villamizar, Diab, and Solorio}]{molina2019overview}
Giovanni Molina, Fahad AlGhamdi, Mahmoud Ghoneim, Abdelati Hawwari, Nicolas
  Rey-Villamizar, Mona Diab, and Thamar Solorio. 2019.
\newblock Overview for the second shared task on language identification in
  code-switched data.
\newblock \emph{arXiv preprint arXiv:1909.13016}.

\bibitem[{Partanen et~al.(2018)Partanen, Lim, Rie{\ss}ler, and
  Poibeau}]{partanen2018dependency}
Niko Partanen, KyungTae Lim, Michael Rie{\ss}ler, and Thierry Poibeau. 2018.
\newblock Dependency parsing of code-switching data with cross-lingual feature
  representations.
\newblock In \emph{International Workshop on Computational Linguistics for
  Uralic Languages}, pages 1--17. ACL.

\bibitem[{Pratapa et~al.(2018)Pratapa, Choudhury, and
  Sitaram}]{pratapa2018word}
Adithya Pratapa, Monojit Choudhury, and Sunayana Sitaram. 2018.
\newblock Word embeddings for code-mixed language processing.
\newblock In \emph{Proceedings of the 2018 conference on empirical methods in
  natural language processing}, pages 3067--3072.

\bibitem[{Raghavi et~al.(2015)Raghavi, Chinnakotla, and
  Shrivastava}]{raghavi2015answer}
Khyathi~Chandu Raghavi, Manoj~Kumar Chinnakotla, and Manish Shrivastava. 2015.
\newblock " answer ka type kya he?" learning to classify questions in
  code-mixed language.
\newblock In \emph{Proceedings of the 24th International Conference on World
  Wide Web}, pages 853--858.

\bibitem[{Roy et~al.(2013)Roy, Choudhury, Majumder, and
  Agarwal}]{roy2013overview}
Rishiraj~Saha Roy, Monojit Choudhury, Prasenjit Majumder, and Komal Agarwal.
  2013.
\newblock Overview of the fire 2013 track on transliterated search.
\newblock In \emph{Post-Proceedings of the 4th and 5th Workshops of the Forum
  for Information Retrieval Evaluation}, pages 1--7.

\bibitem[{Sanh et~al.(2019)Sanh, Debut, Chaumond, and
  Wolf}]{sanh2019distilbert}
Victor Sanh, Lysandre Debut, Julien Chaumond, and Thomas Wolf. 2019.
\newblock Distilbert, a distilled version of bert: smaller, faster, cheaper and
  lighter.
\newblock \emph{arXiv preprint arXiv:1910.01108}.

\bibitem[{Singh et~al.(2018{\natexlab{a}})Singh, Sen, and
  Kumaraguru}]{singh2018twitter}
Kushagra Singh, Indira Sen, and Ponnurangam Kumaraguru. 2018{\natexlab{a}}.
\newblock A twitter corpus for hindi-english code mixed pos tagging.
\newblock In \emph{Proceedings of the Sixth International Workshop on Natural
  Language Processing for Social Media}, pages 12--17.

\bibitem[{Singh and Lefever(2020)}]{singh2020sentiment}
Pranaydeep Singh and Els Lefever. 2020.
\newblock Sentiment analysis for hinglish code-mixed tweets by means of
  cross-lingual word embeddings.
\newblock In \emph{Proceedings of the The 4th Workshop on Computational
  Approaches to Code Switching}, pages 45--51.

\bibitem[{Singh et~al.(2018{\natexlab{b}})Singh, Vijay, Akhtar, and
  Shrivastava}]{singh2018named}
Vinay Singh, Deepanshu Vijay, Syed~Sarfaraz Akhtar, and Manish Shrivastava.
  2018{\natexlab{b}}.
\newblock Named entity recognition for hindi-english code-mixed social media
  text.
\newblock In \emph{Proceedings of the seventh named entities workshop}, pages
  27--35.

\bibitem[{Sitaram et~al.(2019)Sitaram, Chandu, Rallabandi, and
  Black}]{sitaram2019survey}
Sunayana Sitaram, Khyathi~Raghavi Chandu, Sai~Krishna Rallabandi, and Alan~W
  Black. 2019.
\newblock A survey of code-switched speech and language processing.
\newblock \emph{arXiv preprint arXiv:1904.00784}.

\bibitem[{Solorio et~al.(2014)Solorio, Blair, Maharjan, Bethard, Diab, Ghoneim,
  Hawwari, AlGhamdi, Hirschberg, Chang et~al.}]{solorio2014overview}
Thamar Solorio, Elizabeth Blair, Suraj Maharjan, Steven Bethard, Mona Diab,
  Mahmoud Ghoneim, Abdelati Hawwari, Fahad AlGhamdi, Julia Hirschberg, Alison
  Chang, et~al. 2014.
\newblock Overview for the first shared task on language identification in
  code-switched data.
\newblock In \emph{Proceedings of the First Workshop on Computational
  Approaches to Code Switching}, pages 62--72.

\bibitem[{Soto and Hirschberg(2018)}]{soto2018joint}
Victor Soto and Julia Hirschberg. 2018.
\newblock Joint part-of-speech and language id tagging for code-switched data.
\newblock In \emph{Proceedings of the Third Workshop on Computational
  Approaches to Linguistic Code-Switching}, pages 1--10.

\bibitem[{Winata et~al.(2021)Winata, Cahyawijaya, Liu, Lin, Madotto, and
  Fung}]{winata2021multilingual}
Genta~Indra Winata, Samuel Cahyawijaya, Zihan Liu, Zhaojiang Lin, Andrea
  Madotto, and Pascale Fung. 2021.
\newblock Are multilingual models effective in code-switching?
\newblock \emph{arXiv preprint arXiv:2103.13309}.

\bibitem[{Winata et~al.(2018)Winata, Madotto, Wu, and Fung}]{winata2018code}
Genta~Indra Winata, Andrea Madotto, Chien-Sheng Wu, and Pascale Fung. 2018.
\newblock Code-switching language modeling using syntax-aware multi-task
  learning.
\newblock \emph{arXiv preprint arXiv:1805.12070}.

\end{thebibliography}
